\documentclass[10pt,twocolumn,letterpaper]{article}

\usepackage{iccv}
\usepackage{times}
\usepackage{epsfig}
\usepackage{graphicx}
\usepackage{amsmath}
\usepackage{amssymb}
\usepackage{gensymb}

\usepackage[breaklinks=true,bookmarks=false]{hyperref}

\iccvfinalcopy 

\def\iccvPaperID{1901} 

\begin{document}

\title{Faster Than Real-time Facial Alignment: A 3D Spatial Transformer Network Approach in Unconstrained Poses}
\author{Chandrasekhar Bhagavatula, Chenchen Zhu, Khoa Luu, and Marios Savvides\\
Carnegie Mellon University\\
Pittsburgh, PA\\
{\tt\small cbhagava@andrew.cmu.edu, zcckernel@cmu.edu, kluu@andrew.cmu.edu,  msavvid@ri.cmu.edu}
}

\maketitle

\begin{abstract}
Facial alignment involves finding a set of landmark points on an image with a known semantic meaning. However, this semantic meaning of landmark points is often lost in 2D approaches where landmarks are either moved to visible boundaries or ignored as the pose of the face changes. In order to extract consistent alignment points across large poses, the 3D structure of the face must be considered in the alignment step. However, extracting a 3D structure from a single 2D image usually requires alignment in the first place. We present our novel approach to simultaneously extract the 3D shape of the face and the semantically consistent 2D alignment through a 3D Spatial Transformer Network (3DSTN) to model both the camera projection matrix and the warping parameters of a 3D model. By utilizing a generic 3D model and a Thin Plate Spline (TPS) warping function, we are able to generate subject specific 3D shapes without the need for a large 3D shape basis. In addition, our proposed network can be trained in an end-to-end framework on entirely synthetic data from the 300W-LP dataset. Unlike other 3D methods, our approach only requires one pass through the network resulting in a faster than real-time alignment. Evaluations of our model on the Annotated Facial Landmarks in the Wild (AFLW) and AFLW2000-3D datasets show our method achieves state-of-the-art performance over other 3D approaches to alignment.
\end{abstract}

\section{Introduction}

Robust face recognition and analysis are contingent upon accurate localization of facial features.  When modeling faces, the landmark points of interest consist of points that lie along the shape boundaries of facial features, e.g. eyes, lips, mouth, etc. When dealing with face images collected in the wild conditions, facial occlusion of landmarks becomes a common problem for off-angle faces.
Predicting the occlusion state of each landmarking points is one of the challenges due to variations of objects in faces, e.g. beards and mustaches, sunglasses and other noisy objects. Additionally, face images of interest nowadays usually contain off-angle poses, illumination variations, low resolutions, and partial occlusions. 

\begin{figure}[t!]
\centering
\includegraphics[width=0.92\columnwidth]{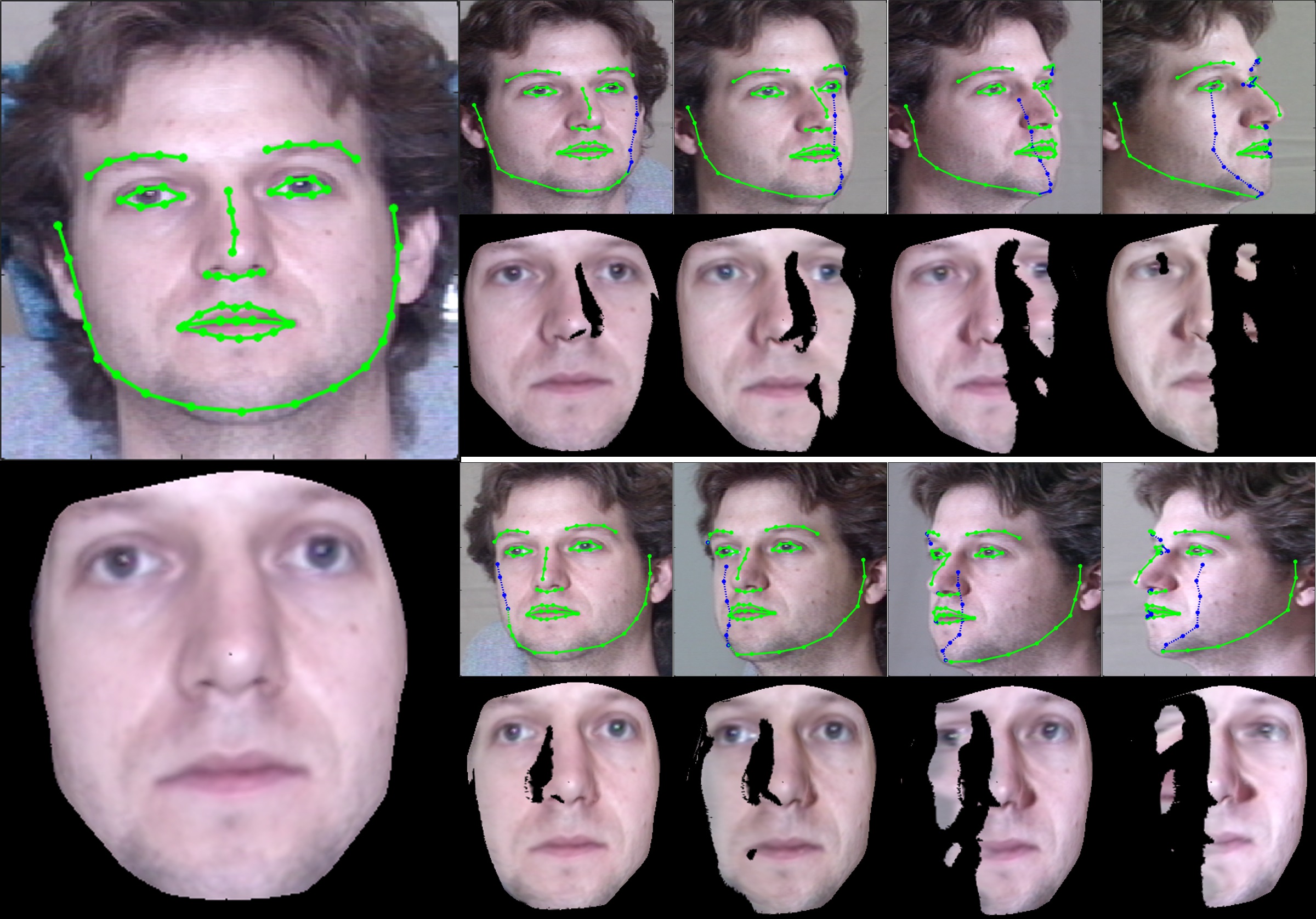}
\caption{A subject from the CMU Multi-PIE dataset \cite{mpie1,mpie2} landmarked and frontalized by our method at various poses. Landmarks found by our model are overlaid in green if they are determined to be a visible landmark and blue if self-occluded. The non-visible regions of the face are determined by the estimated camera center and the estimated 3D shape. \textbf{Best viewed in color.}}
\label{fig:frontalization}
\end{figure}

Many complex factors could affect the appearance of a face image in real-world scenarios and providing tolerance to these factors is the main challenge for researchers. Among these factors, pose is often the most important factor to be dealt with. It is known that as facial pose deviates from a frontal view, most face recognition systems have difficulty in performing robustly. In order to handle a wide range of pose changes, it becomes necessary to utilize 3D structural information of faces. However, many of the existing 3D face modeling schemes ~\cite{Atick96,Blanz02,Wang06} have many drawbacks, such as computation time and complexity. Though these can be mitigated by using depth sensors \cite{Hsieh2015} or by tracking results from frame to frame in video \cite{Saito2016}, this can cause difficulty when they have to be applied in real-world large scale unconstrained face recognition scenarios where video and depth information is not available. The 3D generic elastic model (3D-GEM) approach was proposed as an efficient and reliable 3D modeling method from a single 2D image. Heo \emph{et al.} \cite{jheo_gem_2009,GEM_2011} claim that the depth information of a face is not extremely discriminative when factoring out the 2D spatial location of facial features. In our method, we follow this idea and observe that fairly accurate 3D models can be generated by using a simple mean shape deformed to the input image at a relatively low computational cost compared to other approaches.

\begin{figure}[t!]
\centering
\includegraphics[width=0.93\columnwidth]{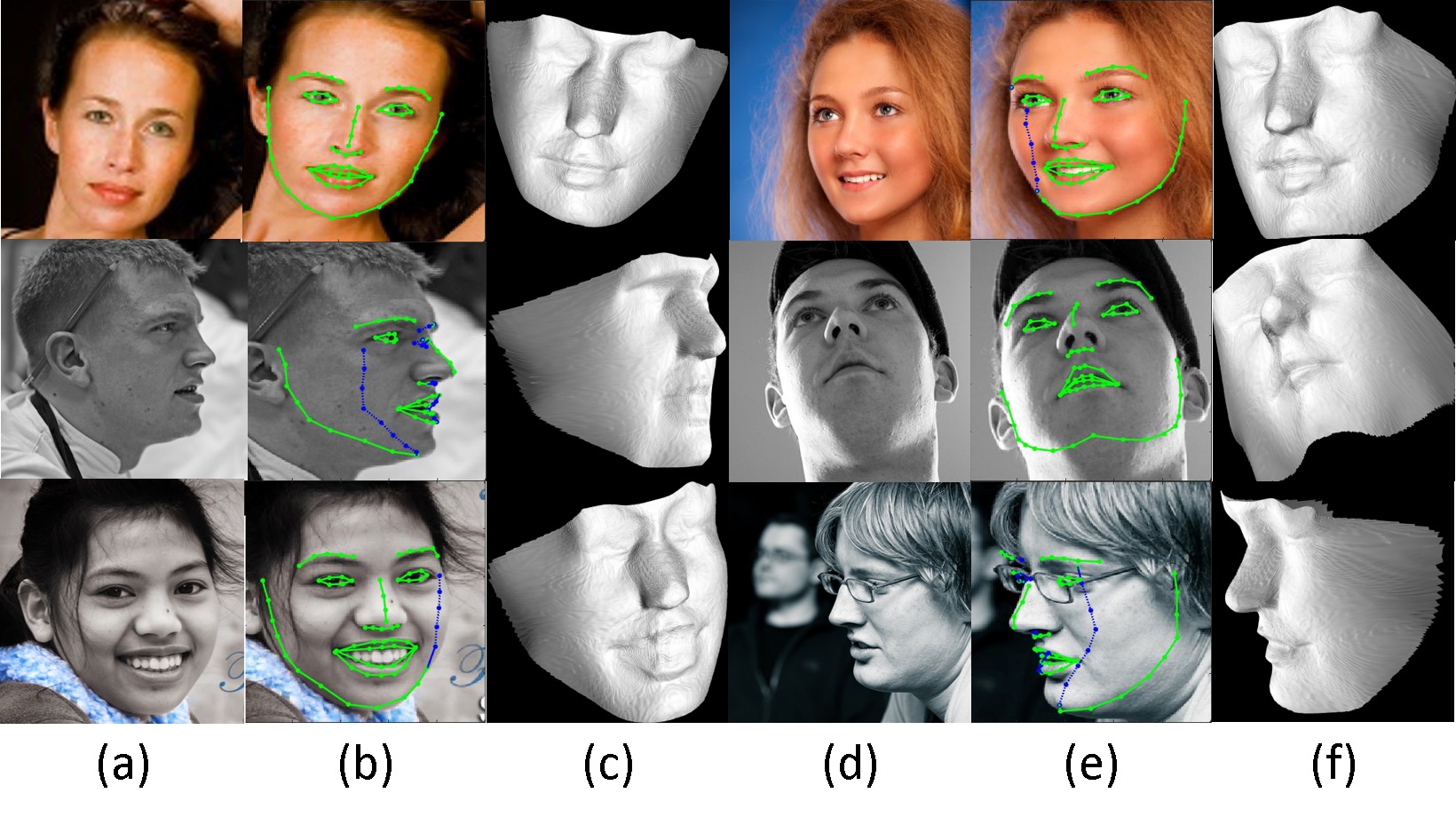}
\caption{(a \& d): Images in the wild from the AFLW dataset. (b \& d): 3D landmarks (green: visible, blue: occluded) estimated from input image. (e \&  f): 3D model generated from input image. \textbf{Best viewed in color.}}
\label{fig:3d_surf}
\end{figure}

\subsection{Our Contributions in this Work}
\textbf{(1)} We take the approach of using a simple mean shape and using a parametric, non-linear warping of that shape through alignment on the image to be able to model any unseen example. A key flaw in many approaches that rely on a 3D Morphable Model (3DMM) is that it needs enough examples of the data to be able to model unseen samples. However, in the case of 3D faces, most datasets are very small.

\textbf{(2)} Our approach is efficiently implemented in an end-to-end deep learning framework allowing for the alignment and 3D modeling tasks to be codependent. This ensures that alignment points are semantically consistent across changing poses of the object which also allows for more consistent 3D model generation and frontalization on images in the wild as shown in Figs. \ref{fig:frontalization} and \ref{fig:3d_surf}. 

\textbf{(3)} Our method only requires a single pass through the network allowing us to achieve faster than real-time processing of images with state-of-the-art performance over other 2D and 3D approaches to alignment.
\section{Related Work}

There have been numerous studies related to face alignment since the first work of Active Shape Models (ASM) \cite{cootes1995active} in 1995. A comprehensive literature review in face alignment is beyond the scope of this work.
In this paper, we mainly focus on
recent Convolutional Neural Network (CNN) approaches to solve the face alignment problem. Especially those methods aimed at using 3D approaches to achieve robust alignment results.

\subsection{Face Alignment Methods}

While Principal Component Analysis (PCA) and its variants \cite{cootes1995active, cootes2001active, cristinacce2008automatic} were successfully used to model the facial shapes and appearances, there have since been many advances in facial alignment.  Landmark locations can be directly predicted by a regression from a learned feature space \cite{cao2014face, dantone2012real, xiong2013supervised}. Xiong et al. \cite{xiong2015global} presented the Global Supervised Descent Method (GSDM) method to solve the problem of 2D face alignment. The objective function in GSDM is divided into multiple regions of similar gradient directions. It then constructs a separate cascaded shape regressor for each region. Yu et al. \cite{yu2013pose} incorporated 3D pose landmarking models with group sparsity to indicate the best landmarks. These kind of methods shows an increase of performance on landmark localization. However, these methods all rely on hand-crafted features. 
Recently, CNN-based methods have achieved good results in facial alignment \cite{Zhu16falp, yu2016deep}. 3DDFA \cite{Zhu16falp} fits a dense 3D face model to the image via CNN and DDN \cite{yu2016deep} proposes a novel cascaded framework incorporating geometric constraints for localizing landmarks in faces and other non-rigid objects. Recently, shape regression has been used in numerous facial landmarking methods \cite{Tzimiropoulos2015, RenICCV2014, Artizzu2013}. 



There are several recent works studying the human head rotations \cite{Cootes2015, Zhu2012}, nonlinear statistical models (\cite{Duong2015BeyondPC}) and 3D shape models \cite{Cao2013, Gu2006}. Nonlinear statistical model approaches are impractical in real-time applications. View-based methods employ a separate model for each viewpoint mode. Traditionally, the modes are specified as part of the algorithm design, and problems can arise at midpoints between models.

\subsection{CNNs for 3D Object Modeling}

While estimating a 3D model from images is not a new problem, the challenging task of modeling objects from a single image has always posed a challenge. This is, of course, due to the ambiguous nature of images where depth information is removed. With the recent success of deep learning and especially CNNs in extracting salient information from images, there have been many explorations into how to best use CNNs for modeling objects in 3 dimensions. Many of these approaches are aimed creating a depth estimation for natural images \cite{Liu16dfm, Bansal16marr, Roy16, Liu15dcnn, Li15dcrf}. While the results on uncontrolled images are impressive, the fact that these models are very general means they tend to suffer when applied to specific objects, such as faces. In fact, many times, the depth estimate for faces in the scene tend to be fairly flat. By limiting the scope of the method, the resulting estimated 3D model can be made much more accurate. Hassner et al. \cite{Hassner15eff} use a 3D model of the face to be able to frontalize faces in unseen images with the end goal of improving face recognition by limiting the variations the matcher has to learn. However, this approach requires landmarks on the input face in the same fashion as other methods \cite{jheo_gem_2009,GEM_2011,Hassner15eff,Masi16pafr,Zhu15hfpe}. 

A 2D approach to landmarking inevitably suffers from the problem of visibility and self-occlusion. As Zhu et al. \cite{Zhu15hfpe} show, the problem of landmark marching, where landmarks tend to move to the visible boundary, can cause issues when estimating 3D models from purely 2D alignment. However, this problem can be alleviated by using a 3D model of the face in the alignment step itself as done in \cite{Jourabloo15pifa, Zhu16falp}. Both of these methods make use of an underlying 3D Morphable Model (3DMM) and try to fit the model to the input image in order to find the 2D landmarks. This of course requires a basis to use and the Basel Face Model (BFM) \cite{bfm09} is a very popular model to use. However, the BFM is only created from a set of 100 male and 100 female scans. As any basis can only recreate combinations of the underlying samples, this can severely limit the capability of these models to fit outlier faces or expressions not seen before. Although there has been recent efforts to generate more accurate 3DMMs \cite{3dmm10k}, neither the data nor the model is available to researchers in the field of biometrics. Therefore, we propose to use a smooth warping function, Thin Plate Splines (TPS) \cite{Bookstein89}, to warp mean shapes to fit the input image and generate new 3D shapes. In this fashion, any new face can be modeled, even if its shape cannot be reconstructed by the BFM.

\section{3D Spatial Transformer Networks}
In order to model how a face truly changes from viewpoint to viewpoint, it is necessary to have both the true 3D model of the subject in the image and the properties of the camera used to capture the image, usually in the form of the camera projection matrix. However, knowledge of the true 3D model and the camera projection matrix are almost always not available. Jaderberg \etal \cite{Jaderberg15}, in their work on Spatial Transformer Networks, use a deep network to estimate the parameters of either an affine transformation or a 2D Thin Plate Spline (TPS) transformation. These parameters are then used to generate a new sampling grid which can then be used to generate the transformed image.

\begin{figure*}[t!]
\centering
\includegraphics[width=1.75\columnwidth]{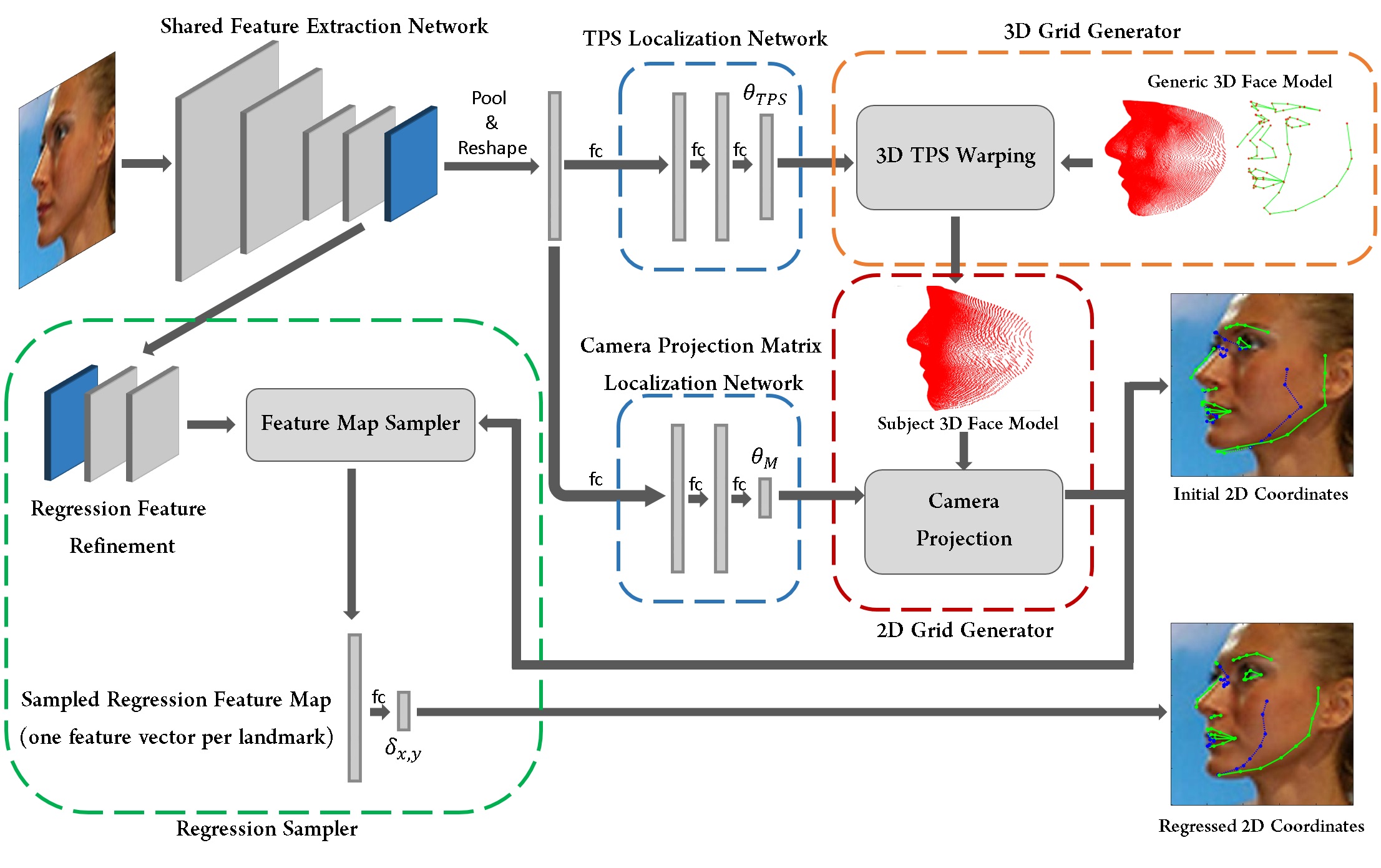}
\caption{Network design of the 3D TPS Spatial Transformer for facial alignment. Because a 3D model and an estimate of the camera position are found in the output of the network, visibility of landmarks can also be determined. Visible landmarks are shown in green while non-visible landmarks are shown in blue.}
\label{fig:3DSPT}
\end{figure*}

We approach finding the unknown camera projection matrix parameters and the parameters needed to generate the 3D model of the head in a similar fashion. Both the camera projection parameters and the warping parameters, a TPS warp in this case, can be estimated from deep features generated from the image using any architecture. The TPS parameters can be used to warp a model of the face to match what the network estimates the true 3D shape is and the camera projection parameters can be used to texture the 3D coordinates from the 2D image. Additionally, the pose of the face can be determined from the camera parameters allowing for a visibility map to be generated for the 3D model. This allows us to only texture vertexes that are visible in the image as opposed to vertexes that are occluded by the face itself. The architecture of our model is shown in Figure \ref{fig:3DSPT}. Sections \ref{sec:CPT},\ref{sec:TPS}, and \ref{sec:WCPT} detail how to create differentiable modules to utilize the camera projection and TPS parameters that are estimated by the deep network to warp and project a 3D model to a 2D image plane for texture sampling.

\subsection{Camera Projection Transformers}
\label{sec:CPT}
In order to be able to perform end-to-end training of a network designed to model 3D transformations of the face, a differentiable module that performs a camera projection must be created. This will be part of the grid generator portion of the Spatial Transformer. Modeling how a 3D point will map to the camera coordinates is expressed by the well known camera projection equation
\begin{equation}
\mathbf{p}_c \widetilde{=} \mathbf{M}\mathbf{p}_w
\label{eqn:proj}
\end{equation}
where $\mathbf{p}_c$ is the homogeneous 2D point in the camera coordinate system, $\mathbf{p}_w$ is the homogeneous 3D point in the world coordinate system, and $\mathbf{M}$ is the 3x4 camera projection matrix. This relationship is only defined up to scale due to the ambiguity of scale present in projective geometry, hence the $\widetilde{=}$ instead of a hard equality. The camera projection matrix has only 11 degrees of freedom since it is only defined up to scale as well. Therefore, this module takes in the 11 parameters estimated by a previous layer as the input in the form of a length 11 vector, $\mathbf{a}$. In order to perform backpropogation on the new grid generator, the derivative of the generated grid with respect to $\mathbf{a}$ must be computed. 

Since Eqn. \ref{eqn:proj} is only defined up to scale, the final output of this module will have to divide out the scale factor. By first rewriting the camera projection matrix as
\begin{equation}
\mathbf{M}=\begin{bmatrix}a_1 & a_2 & a_3 & a_4\\a_5 & a_6 & a_7 & a_8\\a_9 & a_{10} & a_{11} & 1\end{bmatrix}=\begin{bmatrix}\mathbf{m}_1^T\\\mathbf{m}_2^T\\\mathbf{m}_3^T\end{bmatrix}
\end{equation}
where $a_i$ is the $i^{th}$ element of $\mathbf{a}$, the final output of the camera projection module can be written as
\begin{equation}
\mathbf{O}=\begin{bmatrix}x_c\\y_c\end{bmatrix}=\begin{bmatrix}\frac{\mathbf{m}_1^T\mathbf{p}_w}{\mathbf{m}_3^T\mathbf{p}_w}\\
\frac{\mathbf{m}_2^T\mathbf{p}_w}{\mathbf{m}_3^T\mathbf{p}_w}\end{bmatrix}
\label{eqn:proj_output}
\end{equation}
The gradient with respect to each of the rows of $\mathbf{M}$ can be shown to be
\begin{gather}
\frac{\delta\mathbf{O}}{\delta\mathbf{m}_1^T}=\begin{bmatrix}\frac{\mathbf{p}_w^T}{\mathbf{m}_3^T\mathbf{p}_w}\\\mathbf{0}\end{bmatrix} \hspace{.2in} \frac{\delta\mathbf{O}}{\delta\mathbf{m}_2^T}=\begin{bmatrix}\mathbf{0}\\\frac{\mathbf{p}_w^T}{\mathbf{m}_3^T\mathbf{p}_w}\end{bmatrix} \nonumber\\
\frac{\delta\mathbf{O}}{\delta\mathbf{m}_3^T}=\begin{bmatrix}\frac{-\mathbf{p}_w^T\left(\mathbf{m}_1^T\mathbf{p}_w\right)}{\left(\mathbf{m}_3^T\mathbf{p}_w\right)^2}\\\frac{-\mathbf{p}_w^T\left(\mathbf{m}_2^T\mathbf{p}_w\right)}{\left(\mathbf{m}_3^T\mathbf{p}_w\right)^2}\end{bmatrix}
\end{gather}
Using the chain rule, the gradient of the loss of the network with respect to the input can be found as
\begin{equation}
\frac{\delta L}{\delta\mathbf{a}}=
\begin{bmatrix}\left(\frac{\delta L}{\delta\mathbf{O}}\frac{\delta\mathbf{O}}{\delta \mathbf{m}_1^T}\right)^T\\ 
\left(\frac{\delta L}{\delta\mathbf{O}}\frac{\delta\mathbf{O}}{\delta \mathbf{m}_2^T}\right)^T\\
\left(\frac{\delta L}{\delta\mathbf{O}}\frac{\delta\mathbf{O}}{\delta \mathbf{m}_3^T}\right)^T\end{bmatrix}
\end{equation}
Since $\mathbf{M}$ is only defined up to scale, the last element of $\mathbf{M}$ can be defined to be a constant which means that only the first 11 elements of this gradient are used to actually perform the backpropogation on $\mathbf{a}$. Since $\mathbf{M}$ relates many pairs of 2D and 3D points, the gradient is computed for every pair and added together to give the final gradient that is used for updating $\mathbf{a}$.

\subsection{3D Thin Plate Spline Transformers}
\label{sec:TPS}
When modeling the 3D structure of a face, a generic model cannot represent the variety of shapes that might be seen in an image. Therefore, some method of warping a model must be used to allow the method to handle unseen shapes. Thin Plate Spline (TPS) warping has been used by many applications to great effect \cite{Bookstein89,Chui03}. TPS warps have the very dersirable features of providing a closed form of a smooth, parameterized warping given a set of control points and desired destination points. Jaderberg \etal \cite{Jaderberg15} showed how 2D TPS Spatial Transformers could lead to good normalization of nonlinearly transformed input images. Applying a TPS to a 3D set of points follows a very similar process. As in \cite{Jaderberg15}, the TPS parameters would be estimated from a deep network of some sort and passed as input to a 3D grid generator module.

A 3D TPS function is of the form
\begin{equation}
f_{\Delta_x}\left(x,y,z\right)=\begin{bmatrix}b_{1x}\\b_{2x}\\b_{3x}\\b_{4x}\end{bmatrix}^T\begin{bmatrix}1\\x\\y\\z\end{bmatrix}+\sum_{j=1}^n w_{jx}U\left(|\mathbf{c}_j-\left(x,y\right)|\right)
\label{eqn:TPS_3D}
\end{equation}
where $b_{1x}$, $b_{2x}$, $b_{3x}$, $b_{4x}$, and $w_{jx}$ are the parameters of the function, $\mathbf{c}_j$ is the $j^{th}$ control point used in determining the function parameters, and $U(r)=r^2\log{r}$. This function is normally learned by setting up a system of linear equations using the known control points, $\mathbf{c}_j$ and the corresponding points in the warped 3D object. The function finds the change in a single coordinate, the change in the $x$-coordinate in the case of Eqn. \ref{eqn:TPS_3D}. Similarly, one such function is created for each dimension, i.e. $f_{\Delta_x}\left(x,y,z\right)$, $f_{\Delta_y}\left(x,y,z\right)$, and $f_{\Delta_z}\left(x,y,z\right)$. The 3D TPS module would then take in the parameters for all three of these functions as input and output the newly transformed points on a 3D structure as
\begin{equation}
\mathbf{O}=\begin{bmatrix}x^\prime\\y^\prime\\z^\prime\end{bmatrix}=\begin{bmatrix}f_{\Delta_x}\left(x,y,z\right)\\f_{\Delta_y}\left(x,y,z\right)\\f_{\Delta_z}\left(x,y,z\right)\end{bmatrix}+\begin{bmatrix}x\\y\\z\end{bmatrix}
\end{equation}
This means that the 3D TPS module must have all of the 3D vertices of the generic model and the control points on the generic model as fixed parameters specified from the start. This will allow the module to warp the specified model by the warps specified by the TPS parameters.

As in \ref{sec:CPT}, the gradient of the loss with respect to the input parameters must be computed in order to perform backpropogation on this module. As usual, the chain rule can be used to find this by computing the gradient of the output with respect to the input parameters. Since each 3D vertex in the generic model will give one 3D vertex as an output, it is easier to compute the gradient on one of these points, $\mathbf{p}_i=(x_i,y_i,z_i)$, first. This can be shown to be
\begin{equation}
\frac{\delta\mathbf{O}}{\delta \boldsymbol{\theta}_{\Delta_x}}=\begin{bmatrix}1&0&0\\x_i&0&0\\y_i&0&0\\z_i&0&0\\U\left(|\mathbf{c}_1-\left(x_i,y_i,z_i\right)|\right)&0&0\\\vdots&\vdots&\vdots\\U\left(|\mathbf{c}_n-\left(x_i,y_i,z_i\right)|\right)&0&0\end{bmatrix}^T
\end{equation}
where $\boldsymbol{\theta}_{\Delta_x}$ are the parameters of $f_{\Delta_x}$. Similarly, the gradients for $\boldsymbol{\theta}_{\Delta_y}$ and $\boldsymbol{\theta}_{\Delta_z}$ are the same with only the non-zeros values in either the second or third row respectively. The final gradient of the loss with respect to the parameters can be computed as
\begin{equation}
\frac{\delta L}{\delta\boldsymbol{\theta}_{\Delta_x}}=\frac{\delta L}{\delta \mathbf{O}}\frac{\delta \mathbf{O}}{\delta \boldsymbol{\theta}_{\Delta_x}}
\end{equation}
Since this is only for a single point, once again the gradient can be computed for every point and added for each set of parameters to get the final gradient for each set of parameters that can be used to update previous layers of the network.

\subsection{Warped Camera Projection Transformers}
\label{sec:WCPT}
In order to make use of the TPS warped 3D points in the camera projection module of the transformer network, the module must take in as input the warped coordinates. This means that such a module would also have to do backpropogation on the 3D coordinates as well as the camera projection parameters. Since \ref{sec:CPT} already specified how to compute the gradient of the loss with respect to the camera projection parameters, all that is left to do is compute the gradient of the loss with respect to the 3D coordinates in this module.
Taking the derivative of the output in Eqn. \ref{eqn:proj_output} with respect to the 3D point, $\mathbf{p}_w$ results in
\begin{equation}
\frac{\delta \mathbf{O}}{\delta\mathbf{p}_w}=\begin{bmatrix}\frac{\mathbf{m}_1^T}{\mathbf{m}_3^T\mathbf{p}_w}-\frac{\mathbf{m}_1^T\mathbf{p}_w}{\left(\mathbf{m}_3^T\mathbf{p}_w\right)^2}\mathbf{m}_3^T\\
\frac{\mathbf{m}_2^T}{\mathbf{m}_3^T\mathbf{p}_w}-\frac{\mathbf{m}_2^T\mathbf{p}_w}{\left(\mathbf{m}_3^T\mathbf{p}_w\right)^2}\mathbf{m}_3^T\end{bmatrix}
\end{equation}
However, since $\mathbf{p}_w$ is in homogeneous coordinates and only the gradient with respect to the $x$, $y$, and $z$ coordinates are needed, the actual gradient becomes
\begin{equation}
\frac{\delta \mathbf{O}}{\delta\mathbf{p}_w^\prime}=\begin{bmatrix}\frac{\mathbf{m}_1^{\prime T}}{\mathbf{m}_3^T\mathbf{p}_w}-\frac{\mathbf{m}_1^T\mathbf{p}_w}{\left(\mathbf{m}_3^T\mathbf{p}_w\right)^2}\mathbf{m}_3^{\prime T}\\
\frac{\mathbf{m}_2^{\prime T}}{\mathbf{m}_3^T\mathbf{p}_w}-\frac{\mathbf{m}_2^T\mathbf{p}_w}{\left(\mathbf{m}_3^T\mathbf{p}_w\right)^2}\mathbf{m}_3^{\prime T}\end{bmatrix}
\end{equation}
where
\begin{equation}
\mathbf{p}_w^\prime=\begin{bmatrix}x_w\\y_w\\z_w\end{bmatrix} \hspace{.2in} \mathbf{m}_i^\prime=\begin{bmatrix}m_{i_1}\\m_{i_3}\\m_{i_3}\end{bmatrix}
\end{equation}
and $m_{i_j}$ is the $j^{th}$ element of $\mathbf{m}_i$. This gradient is computed for every 3D point independently and used in the chain rule to compute
\begin{equation}
\frac{\delta L}{\delta\mathbf{p}_w}=\frac{\delta L}{\delta \mathbf{O}}\frac{\delta \mathbf{O}}{\delta \mathbf{p}_w}
\end{equation}
which can then be used to perform backpropogation on each $\mathbf{p}_w$.

\subsection{2D Landmark Regression}
In order to further improve the landmark accuracy, we extend our network with a landmark refinement stage. This stage treats the projected 2D coordinates from the previous stage as initial points and estimates the offsets for each point. To extract the feature vector for each point, a $3\times 3$ convolution layer is attached on top of the last convolution layer in the base model, followed by a $1\times 1$ convolution layer for more nonlinearity, resulting in a feature map with $D$ channels. Then each initial point is projected onto this feature map and its $D$-dimensional feature vector is extracted along the channel direction. Notice that the initial points are often not aligned with the grids on the feature map. Therefore, their feature vectors are sampled with bilinear interpolation.

Given the feature vector for each landmark, it goes through a fully-connected (FC) layer to output the offsets, i.e. $\delta_x$ and $\delta_y$. Then the offsets are added to the coordinates of the initial location. For each landmark we use an independent FC layer. We don't share the FC layer for all landmarks because each landmark should have a unique behavior of offsets. For example, the center of the eye may move left after regression whereas the corner of the eye may move right. Also, sometimes two initial landmarks may be projected to the same location due to a certain pose. We want them to move to different locations even when they have the same feature vector.

\subsection{3D Model Regression From 2D Landmarks}

Once the 2D regression is performed, the mapping between the 3D model and the 2D landmarks is broken. While this is not necessarily a problem in the case of sparse facial alignment, if a denser scheme is needed, the entire model would have to be retrained. In order to avoid this, we create a new 3D model that does map to these 2D landmarks by finding a new set of 3D coordinates that project to the new 2D landmarks and warping the 3D model to fit these new points. To find the new 3D coordinates, we need to backproject rays through each of the 2D landmarks through 3D space using the camera projection matrix we have estimated. The equation for the ray of points associated with a given homogeneous 2D point, $\mathbf{p}^i_{2D}$, is defined as
\begin{equation}
\mathbf{p}^{i\prime}_{3D}=\begin{bmatrix}\mathbf{A}^{-1}\mathbf{b}\\1\end{bmatrix}+\lambda \begin{bmatrix}\mathbf{A}^{-1}\mathbf{p}^i_{2D}\\ 0\end{bmatrix}
\end{equation}
where $\mathbf{A}$ and $\mathbf{b}$ are the first three and the last column of the estimated camera projection matrix respectively.

\begin{figure}[t!]
\centering
\includegraphics[width=.9\columnwidth]{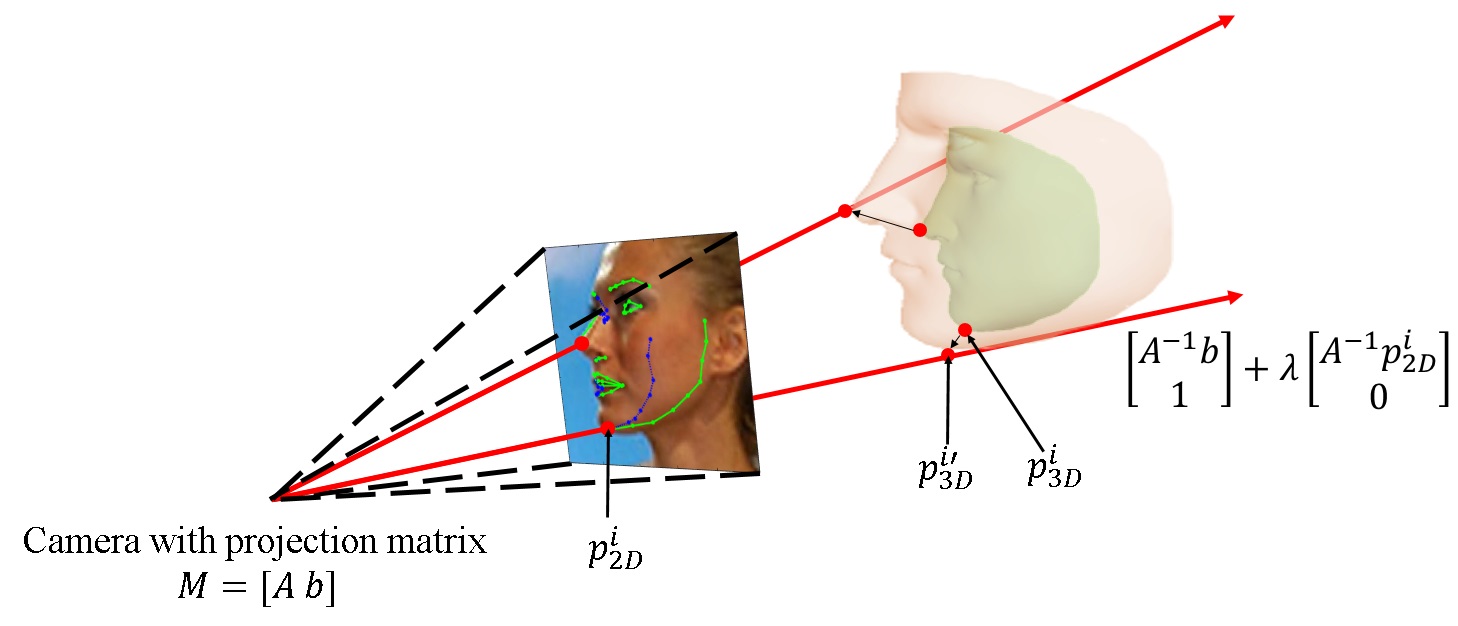}
\caption{Backprojection or rays through image landmarks. The closest points are found for each ray-landmark pair to use as new 3D coordinates for the face model. The original model (green) is warped to fit the new landmarks with a 3D TPS warp resulting in a new face model (red).}
\label{fig:3Dwarp}
\end{figure}

\begin{figure}[t!]
\centering
\includegraphics[width=\columnwidth]{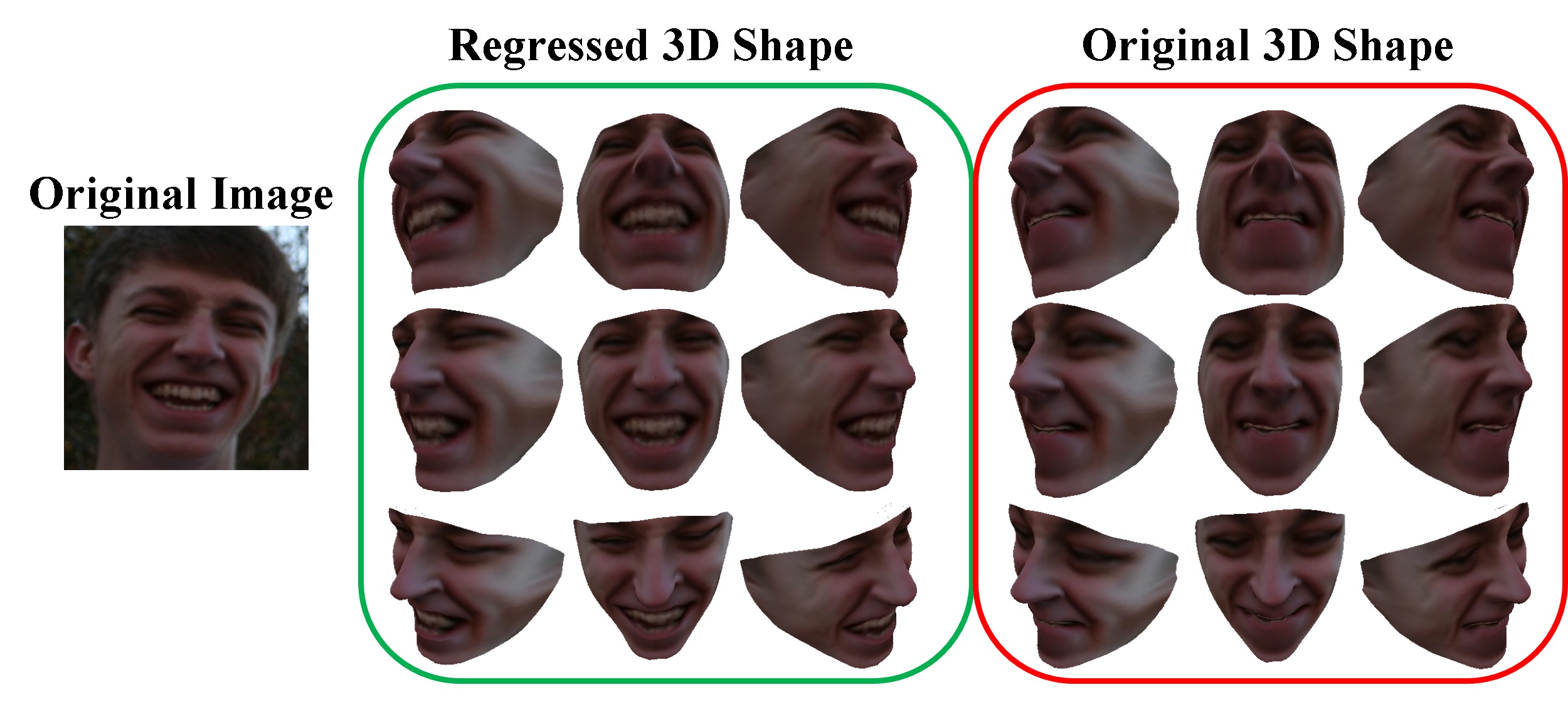}
\caption{3D renderings of input face with a smiling expression. The resulting regressed 3D model (green box) maintains the smile and is very similar to the input image while the same texture applied to the original shape (red box) suffers a small degradation in shape but allows for a non-smiling rendering of the input image.}
\label{fig:3Dexpr}
\end{figure}

These rays represent all possible points in 3D that could project to the determined locations in the image. We then find the closest point, $\mathbf{p}^{i\prime}_{3D}$, on the ray to the original 3D coordinate, $\mathbf{p}^{i}_{3D}$, to use as the new 3D point as shown in Fig. \ref{fig:3Dwarp}. These new correspondences are used to perform a TPS warping of the model. After this warping, the landmark points on the model will project to exactly the regressed 2D landmarks, recovering the mapping between the 3D model and the 2D image. This new model can then be projected onto the image to generate a much more accurate texturing of the 3D model. This same style of warping can be used to move the 3D coordinates anywhere we choose. This means neutralizing out expressions, especially smiles, is very easy to do by using the texture from the regressed 3D shape. While the non-smiling shape will not be as accurate due to the fact that a non-smiling image was not seen, it still gives convincing qualitative results, as seen in Fig. \ref{fig:3Dexpr}, which indicate it may be a worthwhile avenue of exploration for future work, especially in face recognition.

\section{Experiments}

\subsection{Datasets}
\textbf{300W-LP}: The 300W-LP \cite{Zhu16falp} dataset contains 122,450 synthetically generated views of faces from the AFW \cite{afw}, LFPW \cite{lfpw}, HELEN \cite{helen}, and IBUG \cite{ibug} datasets. These images not only contain rotated faces but also attempt to move the background in a convincing fashion, making it a very useful dataset for training 3D approaches to work on real world images.

\textbf{AFLW}: The Annotated Facial Landmarks in the Wild (AFLW) dataset \cite{aflw} is a relatively large dataset for evaluating facial alignment on wild images. It contains approximately 25,000 faces annotated with 21 landmarks with visibility labels. The dataset provides pose estimates so results are grouped into three different pose ranges, $[0\degree, 30\degree]$, $(30\degree, 60\degree]$, and $(60\degree, 90\degree]$. Due to the inconsistency in the bounding boxes in the AFLW dataset, we adopt the use of a face detector first to normalize the scale of the faces. The Multiple Scale Faster Region-based CNN approach \cite{msfrcnn} has shown good results and at a fast speed. We use the recent extension to this work, the Contextual Multi-Scale Region-based CNN (CMS-RCNN) approach \cite{cmsrcnn} to perform the face detection in any experiment where face detection is needed. The CMS-RCNN approach detects 98.8\% (13,865), 95.9\% (5,710), and 86.5\% (3,830) of the faces in the $[0\degree, 30\degree]$, $(30\degree, 60\degree]$, and $(60\degree, 90\degree]$ pose ranges respectively.

\textbf{AFLW2000-3D}: Zhu et al. \cite{Zhu16falp} accurately pointed out how merely evaluating an alignment scheme on the visible landmarks in a dataset can result in artificially low errors. Therefore, a true evaluation of any 3D alignment method must also evaluate alignment on the non-visible landmarks as well. The AFLW2000-3D dataset contains the first 2000 images of the AFLW dataset but with all 68 points defined by the scheme in the CMU MPIE dataset \cite{mpie1,mpie2}. These points were found by aligning the Basel Face Model to the images. While this is a synthetic dataset, meaning the true location of the non-visible landmarks is not known, it is the best one can do when dealing with real images. As these images are from the AFLW dataset, they are also grouped into the same pose ranges.

\begin{figure*}[t!]
\centering
\includegraphics[width=1.75\columnwidth]{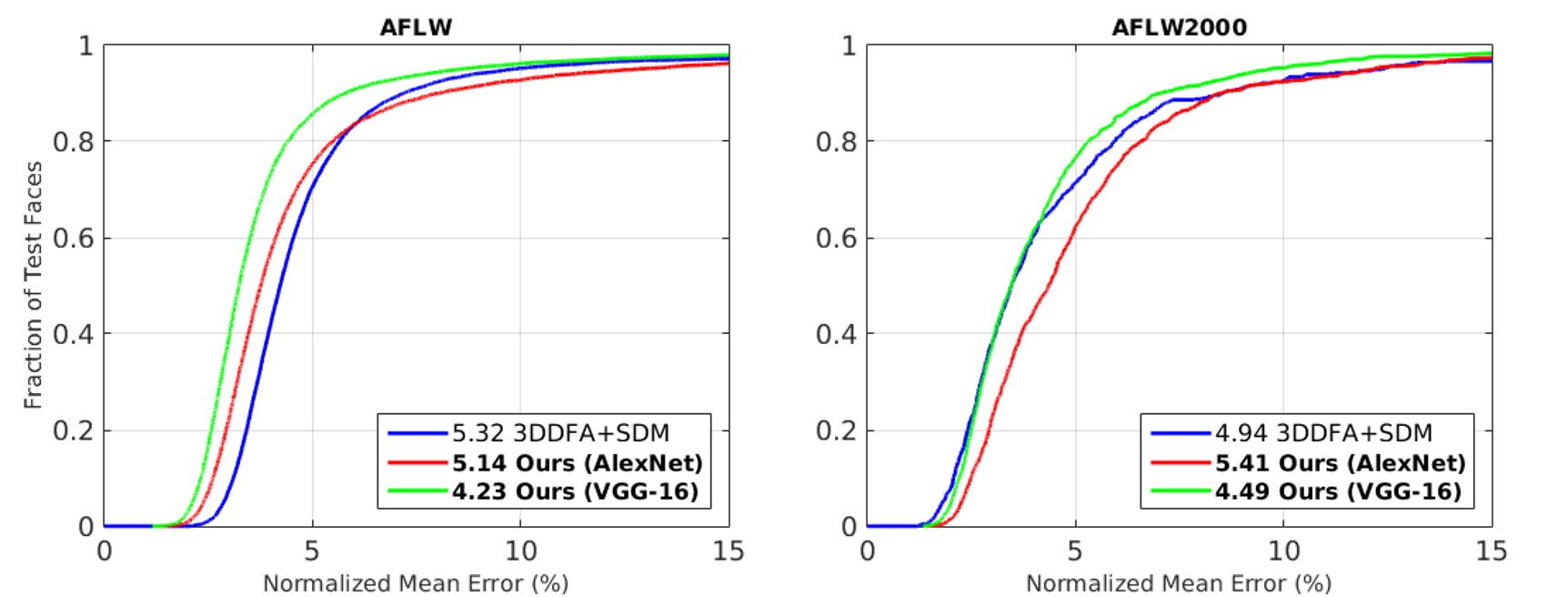}
\caption{CED curves for both the AlexNet (red) and VGG-16 (green) architectures on both the AFLW (left) and AFLW2000-3D (right) dataset. To balance the distributions, we randomly sample 13,209 faces from AFLW and 915 faces from AFLW2000-3D, split evenly among the 3 categories, and compute the CED curve. This is done 10 times and the average of the resulting CED curves are reported. The mean NME\% for each architecture from Table \ref{tab:exp_results} is also reported in the legend.}
\label{fig:ced}
\end{figure*}

\subsection{Implementation Details}
Our network is implemented in the Caffe \cite{jia2014caffe} framework. A new layer is created consisting of the 3D TPS transformation module, the camera projection module and the bilinear sampler module. All modules are differentiable so that the whole network can be trained end-to-end.

We adopt two architectures, AlexNet \cite{krizhevsky2012imagenet} and VGG-16 \cite{simonyan2014very}, as the pre-trained models for our shared feature extraction networks in Fig. \ref{fig:3DSPT}, i.e. we use the convolution layers from the pre-trained models to initialize ours. Since these networks already extract informative low-level features and we do not want to lose this information, we freeze some of the earlier convolution layers and finetune the rest. For the AlexNet architecture, we freeze the first layer while for the VGG-16 architecture, the first 4 layers are frozen.

The 2D landmark regression is implemented by attaching additional layers on top of the last convolution layer. With $N$ landmarks to regress, we need $N$ FC layers to compute the offsets for each individual landmark. While it's possible to setup $N$ individual FC layers, here we implement this by adding one Scaling layer followed by a Reduction layer and Bias layer. During training only the new layers are updated and all previous layers are frozen.

\subsection{Training on 300W-LP}
When training our model, we train on the AFW, HELEN, and LFPW subsets of the 300W-LP dataset and use the IBUG portion as a validation set. All sets are normalized using the bounding boxes from the CMS-RCNN detector by reshaping the detected faces to 250 x 250 pixels.
For the AlexNet architecture, we train for 100,000 iterations with a batch size of 50. The initial learning rate is set to 0.001 and drops by a factor of 2 after 50,000 iterations. When training the landmark regression, the initial learning rate is 0.01 and drops by a factor of 10 every 40,000 iterations. 
For the VGG-16 architecture, we train for 200,000 iterations with a batch size of 25. The initial learning rate is set to 0.001 and drops by a factor of 2 after 100,000 iterations. When training the landmark regression, the initial learning rate is 0.01 and drops by a factor of 10 every 70,000 iterations. 
The momentum for all experiments is set to 0.9. Euclidean loss is applied to 3D vertexes, 2D projected landmarks and 2D regressed landmarks.

\begin{table}[t!]
\centering
\caption{Alignment accuracy for both the AlexNet (AN) and VGG-16 (VGG) models. (LR: landmark regression)}
\label{tab:ablation}
\begin{tabular}{|c||c|c|c|c|c|}
\hline
 & \multicolumn{5}{c|}{AFLW Dataset (21 pts)} \\ \hline
 & $[0, 30]$ & $(30, 60]$ & $(60, 90]$ & mean & std \\ \hline \hline
AN & 4.88 & 5.55 & 7.10 & 5.84 & 1.14 \\ \hline
AN+LR & 4.00 & 4.48 & 5.89 & 4.79 & 0.98 \\ \hline \hline
VGG & 4.15 & 4.64 & 5.96 & 4.92 & 0.94 \\ \hline
VGG+LR & 3.46 & 3.78 & 4.77 & 4.00 & 0.69 \\ \hline
\end{tabular}
\end{table}

\begin{table*}[t!]
\centering
\caption{The NME(\%) of face alignment results on AFLW and AFLW2000-3D. The best two numbers in each category are shown in bold.}
\label{tab:exp_results}
\begin{tabular}{|c||c|c|c|c|c||c|c|c|c|c|}
\hline
 & \multicolumn{5}{c||}{AFLW Dataset (21 pts)} & \multicolumn{5}{c|}{AFLW 2000-3D Dataset (68 pts)} \\ \hline
Method & $[0, 30]$ & $(30, 60]$ & $(60, 90]$ & mean & std & $[0, 30]$ & $(30, 60]$ & $(60, 90]$ & mean & std \\ \hline
CDM & 8.15 & 13.02 & 16.17 & 12.44 & 4.04 & - & - & - & - & - \\ \hline
RCPR & 5.43 & 6.58 & 11.53 & 7.85 & 3.24 & 4.26 & 5.96 & 13.18 & 7.80 & 4.74\\ \hline
ESR & 5.66 & 7.12 & 11.94 & 8.24 & 3.29 & 4.60 & 6.70 & 12.67 & 7.99 & 4.19 \\ \hline
SDM & 4.75 & 5.55 & 9.34 & 6.55 & 2.45 & 3.67 & 4.94 & 9.76 & 6.12 & 3.21 \\ \hline
3DDFA & 5.00 & 5.06 & 6.74 & 5.60 & 0.99 & 3.78 & 4.54 & 7.93 & 5.42 & 2.21 \\ \hline
3DDFA+SDM & 4.75 & 4.83 & \textbf{6.38} & 5.32 & \textbf{0.92} & \textbf{3.43} & \textbf{4.24} & \textbf{7.17} & \textbf{4.94} & 1.97 \\
\hline
\textbf{Ours (AlexNet)} &\textbf{4.11} & \textbf{4.69} & 6.61 & \textbf{5.14} & 1.31 & 3.71 & 5.33 & 7.19 & 5.41 & \textbf{1.74} \\ \hline
\textbf{Ours (VGG-16)} & \textbf{3.55} & \textbf{3.92} & \textbf{5.21} & \textbf{4.23} & \textbf{0.87} & \textbf{3.15} & \textbf{4.33} & \textbf{5.98} & \textbf{4.49} & \textbf{1.42}  \\ \hline
\end{tabular}
\end{table*}

\subsection{Ablation Experiments}
To investigate the effect of each component in our network, we conduct two ablation studies. All the models in these experiments are trained on the same 300W\_LP dataset and tested on the detected images in AFLW. We first test the effect of the different pre-trained models. We fine-tune our network from the AlexNet and VGG-16 models pre-trained on the ImageNet dataset and evaluate the landmark accuracy before the regression step. The VGG-16 model outperforms the AlexNet model in all three pose ranges on the AFLW detected set as shown in Table \ref{tab:ablation}. This seems to indicate that a good base model is important for the parameter estimation portion of the network. Second, we evaluate the effect of landmark regression stage. We compare the errors between the regressed and projected landmarks. Table \ref{tab:ablation} shows that the landmark regression step greatly helps to improve the accuracy.

\subsection{Comparison Experiments}
\textbf{AFLW}: Since the CMS-RCNN approach may only detect the easier to landmark faces, we use the provided bounding box anytime the face is not detected by the detector. Due to the inconsistency between the two bounding box schemes, faces are not always normalized properly. However, we feel this is the only way to get a fair comparison to other methods without artificially making the dataset easier by only evaluating on detected faces. We compare against baseline methods used by \cite{Zhu16falp} on the same dataset, namely Cascaded Deformable Shape Models (CDM) \cite{cdm}, Robust Cascaded Pose Regression (RCPR) \cite{rcpr}, Explicit Shape Regression (ESR) \cite{esr}, SDM \cite{xiong2013supervised} and 3DDFA \cite{Zhu16falp}. All methods except for CDM were retrained on the 300W-LP dataset. The Normalized Mean Error (NME) is computed by averaging the error of the visible landmarks and normalizing it by the square root of the bounding box size ($h$ x $w$) provided in the dataset. Table \ref{tab:exp_results} clearly shows that our model using the VGG-16 architecture has achieved better accuracy in all pose ranges, especially the $(60\degree, 90\degree]$ category, and has achieved a smaller standard deviation in the error. This means that not only are the landmarks more accurate, they are more consistent than the other methods.

\textbf{AFLW2000-3D}: The baseline methods were evaluated using the bounding box of the 68 landmarks so we retrained our models using the same bounding box on the training data. Generating these is trivial due to the 3D models. The NME is computed using the bounding box size. Here we see that though 3DDFA+SDM performs well, the VGG-16 architecture of our model still performs best in both the $[0\degree, 30\degree]$ and $(60\degree, 90\degree]$ ranges. While the VGG-16 model is only second best in the $(30\degree, 60\degree]$ range by a small amount, the improvement in $(60\degree, 90\degree]$ means that, once again, our method generates more accurate and more consistent landmarks, even in a 3D sense. Cumulative Error Distribution (CED) curves are reported for both architectures on both datasets in Fig. \ref{fig:ced}.


\subsection{Running Speed}
In order to evaluate the speed of our method, we evaluate the models on a random subset of 1200 faces from the AFLW subset split evenly into the $[0\degree, 30\degree]$, $(30\degree, 60\degree]$, and $(60\degree, 90\degree]$ pose ranges. The images are processed one at a time to avoid any benefit from batch processing. The models are evaluated on a 3.40 GHz Intel Core i7-6700 CPU and an NVIDIA GeForce GTX TITAN X GPU. Our AlexNet trained model takes a total of 7.064 seconds to landmark the 1200 faces for an average of 0.0059 seconds per image or approximately 170 faces per second. The deeper and more accurate VGG-16 model landmarks the 1200 faces in 22.765 seconds for an average of 0.0190 seconds or approximately 52 faces per second. In comparison, the 3DDFA approach \cite{Zhu16falp} takes 75.72 ms (3 iterations at 25.24 ms per iteration as specified in \cite{Zhu16falp}) with $2/3$ of the time being used to process data on the CPU. 

\section{Conclusions}
In this paper we propose a method using 3D Spatial Transformer Networks with TPS warping to generate both a 3D model of the face and accurate 2D landmarks across large pose variation. The limited data used in the generation of a 3DMM can mean that unseen face shapes cannot be modeled. By using a TPS warp, any potential face can be modeled through a regression of 2D landmarks, of which there is much more data available. We have shown how this approach leads to more accurate and consistent landmarks over other 2D and 3D methods.

{\small
\bibliographystyle{ieee}
\bibliography{egbib}
}

\end{document}